%% file: main.tex
\pdfoutput=1

\documentclass[11pt]{article}
\usepackage{acl}
\usepackage{xurl}
\usepackage{tabularx}
\usepackage{booktabs}
\usepackage{lipsum}
\usepackage{float}
\usepackage{caption}
\usepackage{booktabs}
\usepackage{amsmath}
\usepackage{float}
\usepackage{times}
\usepackage{latexsym}
\usepackage{multirow}
\usepackage{graphicx}
\usepackage[T1]{fontenc}

\usepackage[utf8]{inputenc}

\usepackage{microtype}

\newcommand{\abr}[1]{\textsc{#1}}

\newcommand{\empspeech}{\abr{EmpSpeech}}

\definecolor{lightblue}{RGB}{181, 179, 242}
\definecolor{lightred}{RGB}{245, 210, 208}
\definecolor{lightgreen}{RGB}{188,245,188}
\definecolor{lightyellow}{RGB}{255,255,186}
\definecolor{darkred}{RGB}{235, 64, 52}
\definecolor{darkgreen}{RGB}{91, 143, 94}
\definecolor{darkblue}{RGB}{68, 63, 204}
\definecolor{darkgold}{RGB}{145, 136, 12}
\title{Mixed Signals: Understanding Model Disagreement in\\Multimodal Empathy Detection}

\author{Maya Srikanth \\
    Columbia University \\
    \textsf{ms6198@columbia.edu} \\\And
    Run Chen \\
    Columbia University \\
    \textsf{runchen@cs.columbia.edu} 
 \\\And
   Julia Hirschberg \\
   Columbia University \\
   \textsf{julia@cs.columbia.edu}
   }

\begin{document}
\maketitle
\begin{abstract}
\input{sections/00-abstract}
\end{abstract}

\section{Introduction}
\input{sections/01-introduction}

\section{Related Work}
\input{sections/02-background}
\section{Experiment 1: Identifying Complex Examples from Modality Disagreement}
\input{sections/03-experiment-1}

\section{Experiment 2: Characterizing Complex Examples}
\input{sections/04-experiment-2}

\section{Discussion and Conclusion}
\input{sections/05-conclusion}


\section*{Limitations}
\input{sections/06-limitations}
\section*{Ethics Statement}
\input{sections/07-ethicalconsiderations}

\section*{Acknowledgments}
\input{sections/08-ack}

\bibliography{custom}
\clearpage
\appendix
\input{sections/appendix}
\end{document}

%% file: sections/00-abstract.tex
Multimodal models play a key role in empathy detection, but their performance can suffer when modalities provide conflicting cues. 
To understand these failures, we examine cases where unimodal and multimodal predictions diverge. 
Using fine-tuned models for text, audio, and video, along with a gated fusion model, we find that such disagreements often reflect underlying ambiguity, as evidenced by annotator uncertainty. 
Our analysis shows that dominant signals in one modality can mislead fusion when unsupported by others. 
We also observe that humans, like models, do not consistently benefit from multimodal input. 
These insights position disagreement as a useful diagnostic signal for identifying challenging examples and improving the robustness of empathy systems.

%% file: sections/01-introduction.tex
Empathy recognition in human communication is a nuanced and multifaceted task and a core component of socially intelligent systems~\cite{fung}. 
Commonly defined as the capacity to understand others and share their emotional experiences, empathy encompasses both cognitive perspective-taking and affective resonance~\cite{baumeister}.
In human interactions, language, speech, and visual cues jointly convey emotional intent~\cite{holler}.
For example, a speaker’s verbal message may appear neutral, yet their vocal prosody or facial expressions may signal warmth or concern. 
It is then the listener's responsibility to draw inferences about meaning based on a \textit{combination} of these signals. 

For AI systems, effectively interpreting these multimodal signals requires not only accurate unimodal representations but also robust integration of potentially conflicting information across modalities.
Despite recent advances in multimodal emotion recognition~\cite{jabeen}, empathy recognition remains particularly complex, as
unlike discrete emotions such as anger or joy, 
empathy often arises from subtle contextual cues that may not align across modalities~\cite{hasan}. 
For example, a neutral utterance might be perceived as warm or concerned when accompanied by a sympathetic tone or expression.
\begin{figure}[t]
\centering
\includegraphics[width=0.9\linewidth]{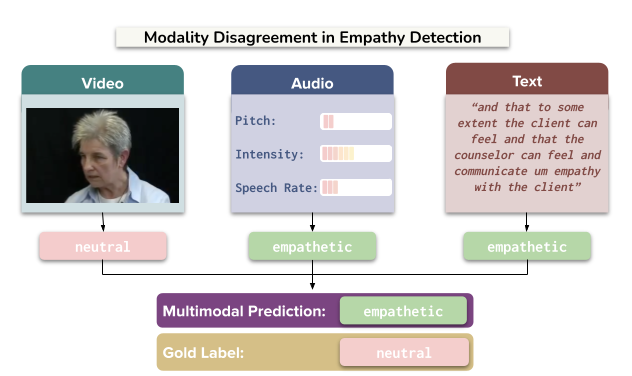}
\caption{Given classifications provided by a single modality, we identify cases where integrating additional modalities leads to a different prediction. We analyze these differences to understand when and why they occur.
}

\label{fig:editing}
\vspace{-1em}
\end{figure}

Our work investigates some of the complexities of multimodal empathy detection by examining instances of disagreement between multimodal models and their unimodal counterparts. 
In parallel, humans annotate unimodal and multimodal examples in our dataset for the presence of empathy.
Our analyses reveal that instances of multimodal and unimodal model disagreement often correspond to examples that are difficult for human annotators as well, highlighting examples that are particularly challenging, ambiguous, or nuanced.
By linking multimodal and unimodal \textit{model} disagreement to \textit{human} disagreement, we offer new insight into the limitations of current empathy modeling and highlight the value of disagreement-based analysis in socially grounded language tasks.

%% file: sections/02-background.tex
\paragraph{Empathy Modeling.} Early computational work on empathy has focused on generating emotionally relevant textual responses~\cite{rashkin, li}, but these approaches are inherently limited by the absence of non-verbal cues, which are critical to empathic understanding.
Recent datasets such as \abr{EmpathicStories++}~\citep{shen}, \abr{medic}~\citep{zhu}, \abr{emmi}~\citep{galland} and \citet{chen} address this limitation by incorporating context, speech, and facial expressions, enabling more comprehensive modeling of empathy.
These resources have motivated frameworks such as \abr{pegs} \citep{zhang}, \abr{EmoKnob} \cite{chen-etal-2024-emoknob} and \abr{synthempathy} \cite{chen2025synthempathyscalableempathycorpus}, which further extend multimodal empathetic generation by leveraging large language models (LLMs) and large audio-language models.
Despite these advances, empathy still remains difficult to model due to its reliance on subtle, often conflicting signals across modalities.
Prior work has largely focused on improving multimodal fusion strategies under the assumption that modalities are complementary \cite{zadeh-etal-2017-tensor, tsai-etal-2019-multimodal}, but has paid less attention to when fusion may fail or introduce noise.
\paragraph{Dataset Difficulty.} 
Complementary lines of work have investigated data difficulty and model disagreement as tools for understanding model behavior.
\citet{swayamdipta} propose \textit{dataset cartography}, a method to identify hard or ambiguous training samples, showing how difficulty-aware instance selection improves benchmarking and reveals mislabeled or trivial examples.
%
%
\citet{saha} demonstrate that difficult instances are also harder for both humans and models to explain, and \citet{wang}'s Learning-From-Disagreement (LFD) framework underscores the importance of examining disagreements between models to gain deeper, actionable insights into their behaviors.
Although ambiguity is intrinsic to empathy modeling, disagreement‐based diagnostics do remain underexplored.
As such, we leverage modality disagreement to flag difficult examples that both mislead fusion models and elicit annotator uncertainty.

%% file: sections/03-experiment-1.tex
\begin{figure*}[t]
\centering
\includegraphics[scale=0.40]{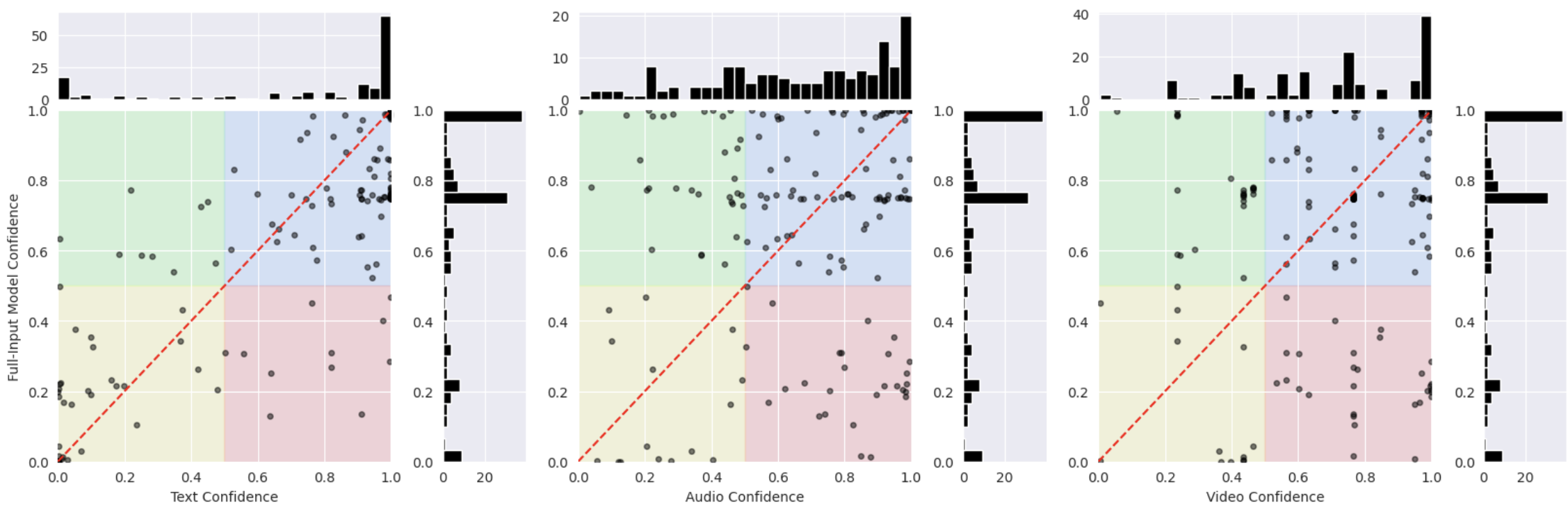}
\caption{Comparing predictions between unimodal (text, audio, video) and multimodal models. We highlight regions where model predictions \textit{agree} (blue and yellow quadrants) and disagree (red and green quadrants).}
\label{fig:conf}
\end{figure*}

Disagreement between models trained on different modalities can reveal challenging, nuanced, or ambiguous examples. 
Here, we identify and analyze such cases of disagreement in binary empathy detection using a multimodal English empathy speech dataset collected from Youtube~\cite{chen} (referred to as \empspeech) consisting of 1,718 manually annotated English speech segments labeled as empathetic or neutral (Appendix~\ref{appendix:data}). 

\paragraph{Experimental Setup.} 
Examples in \empspeech~include video segments spanning three modalities: text (transcript), audio (speech), and video.
The task is to predict whether the input contains empathetic (1) or neutral (0) speech.

We finetune two models per modality on the training set from \abr{EmpSpeech}: \abr{RoBERTa}~\cite{roberta} and \abr{DeBERTa}~\cite{deberta} for text, \abr{HuBERT}~\cite{hubert} and \abr{Wav2Vec2}~\cite{wav2vec2} for audio, and \abr{VideoMAE}~\cite{videomae} and \abr{TimeSFormer}~\cite{timesformer} for video (Appendix~\ref{appendix:uni-model-details}).\footnote{The hidden layer dimensions of all models we consider are similar.} 
Then, we extract 768-dimensional embeddings from each best-performing unimodal model (\abr{RoBERTa}, \abr{HuBERT}, and \abr{VideoMAE}; Table~\ref{tab:modality-performance}) to train a multimodal fusion model that projects all three modalities into a shared latent space (Appendix~\ref{appendix:fusion-model-details}). 
Each modality embedding passes through an independent sigmoid gate that adaptively scales its contribution before fusion. 
The gated embeddings are then passed through an additive attention layer: each is projected into a shared attention space and scored against a learned attention vector. 
These scores are normalized across modalities to compute a weighted sum that forms the fused representation.
\input{tables/mod-perf}
\paragraph{Results.}
We evaluate all models (unimodal and multimodal) on the test split of \empspeech~to identify \textit{disagreements}, or examples where two models with varying input modalities assign \textit{different} labels, highlighting those cases where different modalities may carry ambiguous, conflicting, or modality-specific signals.

Text shows the highest disagreement with audio and video (Table~\ref{tab:disagreement-matrix}), while audio and video align more closely. 
This difference likely reflects shared nonverbal cues such as prosody and facial expression. 
The fusion model’s minimal disagreement with text suggests a bias toward verbal content, possibly mirroring the annotators’ own reliance on textual signals.

Figure~\ref{fig:conf} visualizes disagreement regions between each unimodal model and the fusion model. 
We plot unimodal confidence (x-axis) against fusion confidence (y-axis) in the correct label; hence confidence greater than 0.5 results in a correct prediction.
This yields four quadrants: \colorbox{lightgreen}{\textcolor{darkgreen}{green}}  (multimodal correct, unimodal incorrect), \colorbox{lightred}{\textcolor{darkred}{red}} (multimodal incorrect, unimodal correct), \colorbox{lightblue}{\textcolor{darkblue}{blue}} (both correct), and \colorbox{lightyellow}{\textcolor{darkgold}{yellow}} (both incorrect). 
Red and green quadrants are disagreement regions which we explore to identify complex examples.
\input{tables/disagreement-matrix}

\input{tables/audio-ttests-small}

\subsection{Modality-Based Feature Analysis}

To better understand examples in disagreement regions, we extract and analyze modality-based human interpretable features.

\paragraph{Audio.} We extract twelve prosodic and paralinguistic features from audio signals: nine low-level acoustic features using \abr{Praat}~\cite{boersma} and \abr{Parselmouth}~\cite{jadoul}, and three high-level affective dimensions: valence, arousal, and dominance using a finetuned \abr{Wav2Vec2}~\cite{wagner} model. 
We compare feature distributions using t-tests for examples in disagreement quadrants (\colorbox{lightred}{\textcolor{darkred}{red}} and \colorbox{lightgreen}{\textcolor{darkgreen}{green}}) compared to those in the \colorbox{lightblue}{\textcolor{darkblue}{blue}} quadrant, signifying non-ambiguous, easy examples. 
Blue examples have several significantly elevated pitch-related values than red examples (Table~\ref{tab:ttest-combined}),
suggesting that stronger prosodic fluctuations are frequently corroborated by other modalities. 
Examples in the green quadrant show significantly higher \emph{Max Intensity} than in blue, potentially reflecting the role of volume-based emphasis in aiding unimodal predictions. 
Both red and green examples exhibit significantly lower arousal than blue examples, suggesting that these less-aroused, subtler examples lack sufficient affective intensity, which misleads both unimodal and multimodal models.
\input{tables/au-ttests-small}
\paragraph{Video.} We examine facial action unit (AU) activations~\cite{tadas} from video. 
AU04 (Brow Lowerer), AU12 (Lip Corner Puller), and AU05 (Upper Lid Raiser) show significant differences across example types, revealing how specific facial expressions contribute to perceptual ambiguity (Table~\ref{tab:au-ttests-small}).
AU04 is more active in red examples than blue, indicating that, despite its visually strong presence, its signal conflicts with other modalities. 
In contrast, AU12, which is associated with positive affect, and AU05, which is linked to attentiveness~\citep{friesen}, both show greater activation in blue examples than in red and green, respectively, suggesting that these expressions may serve as clearer cues that are more consistently interpreted across modalities. 
Our findings indicate that fine-grained facial signals may contribute to perceptual complexity in the visual stream.

\paragraph{Text.} Visualizing \abr{umap}~\cite{umap} projections of text embeddings (Figure \ref{fig:umap}) reveals that examples in disagreement regions (red and green) cluster along the boundary between consistently correct (blue) and consistently incorrect (yellow) examples. 
Rather than forming isolated clusters, disagreement examples occupy transition zones in the embedding space: areas where semantic cues are weak. 
This underscores our finding that red and green examples are ambiguous and confirms modality disagreement as a reliable marker of challenging examples in empathy detection.
\input{tables/umap}
\input{tables/entropy}
\subsection{Uncertainty Analysis}
To ensure that the patterns observed in the disagreement quadrants are not simply a byproduct of model uncertainty, we compute the mean predictive entropy from the fusion model’s posterior for examples of each quadrant (Table \ref{tab:entropy}). 

We observe a pronounced divergence in uncertainty: the disagreement quadrants (red and green) have a substantially higher mean predictive entropy than those of the combined agreement quadrants (blue and yellow). Independent-samples \textit{t}-tests at $\alpha = 0.05$ confirm that the difference is statistically significant, with disagreement quadrants showing a higher mean predictive entropy than agreement quadrants ($p = 0.001$). This disparity indicates that model disagreement often co-occurs with high uncertainty, suggesting that examples in the red and green quadrants are both challenging and inherently ambiguous due to conflicting modality signals.

%% file: tables/mod-perf.tex
\begin{table}[t]
\small
\centering
\begin{tabular}{cccc}
\toprule
\textbf{Modality} & \textbf{Model}     & \textbf{Accuracy}&\textbf{F1} \\
\midrule
\multirow{2}{*}{Text}  
    & \textbf{RoBERTa}        & \textbf{0.75\textpm  0.02} & \textbf{0.73 \textpm 0.02} \\
    & DeBERTa         & 0.69\textpm0.02 & 0.68\textpm0.02\\
\midrule
\multirow{2}{*}{Audio} 
    & \textbf{HuBERT}          & \textbf{0.72\textpm0.01} &\textbf{0.71\textpm0.01} \\
    & Wav2Vec2        & 0.68\textpm0.01 &  0.63\textpm0.02 \\
\midrule
\multirow{2}{*}{Video} 
    & \textbf{VideoMAE}        & \textbf{0.77\textpm0.02} &\textbf{0.77\textpm0.02}\\
    & TimesFormer     & 0.64\textpm0.02 &0.62\textpm0.02 \\
\midrule
\multicolumn{2}{l}{\textbf{Fusion (All Modalities)}} & \textbf{0.76\textpm0.02} &\textbf{0.72\textpm0.02} \\
\bottomrule
\end{tabular}
\caption{Fine-tuned model performance by modality on empathy classification (mean ± std over five runs).
}
\label{tab:modality-performance}
\vspace{-14pt}
\end{table}

%% file: tables/disagreement-matrix.tex
\begin{table}[t]
\small
\centering
\begin{tabular}{cccc}
\toprule
\textbf{Modality} & \textbf{Text} & \textbf{Audio} & \textbf{Video} \\
\midrule
\textbf{Text}   & --     & 0.338  & 0.318 \\
\textbf{Audio}  & 0.338  & --     & 0.253 \\
\textbf{Video}  & 0.318  & 0.253  & --     \\
\textbf{Full}  & 0.214  & 0.383  & 0.331     \\
\bottomrule
\end{tabular}
\caption{Pairwise disagreement rates among unimodal models and the fusion model, computed as the fraction of test examples with differing predictions.}
\label{tab:disagreement-matrix}

\end{table}

%% file: tables/audio-ttests-small.tex
\begin{table}[t]
\tiny
\centering
\resizebox{\columnwidth}{!}{
\begin{tabular}{lcccrr}
\toprule
\textbf{Feature} & \multicolumn{2}{c}{\textbf{Red vs. Blue}} & \multicolumn{2}{c}{\textbf{Green vs. Blue}} \\
\cmidrule(lr){2-3} \cmidrule(lr){4-5}
& p-value & Direction & p-value & Direction \\
\midrule
\textbf{valence}        & \textbf{0.0047} & \textbf{$\mu_{\text{blue}} > \mu_{\text{red}}$}   & 0.5166 & $\mu_{\text{green}} > \mu_{\text{blue}}$ \\
\textbf{arousal}        & \textbf{0.0065} & \textbf{$\mu_{\text{blue}} > \mu_{\text{red}}$}   & \textbf{0.0136} & \textbf{$\mu_{\text{blue}} > \mu_{\text{green}}$} \\
\textbf{Mean Pitch}     & \textbf{0.0100} & \textbf{$\mu_{\text{blue}} > \mu_{\text{red}}$}   & \textbf{0.0001} & \textbf{$\mu_{\text{blue}} > \mu_{\text{green}}$} \\
\textbf{dominance}      & \textbf{0.0108} & \textbf{$\mu_{\text{blue}} > \mu_{\text{red}}$}   & 0.0667 & $\mu_{\text{blue}} > \mu_{\text{green}}$ \\
\textbf{Min Pitch}      & \textbf{0.0333} & \textbf{$\mu_{\text{blue}} > \mu_{\text{red}}$}   & \textbf{0.0001} & \textbf{$\mu_{\text{blue}} > \mu_{\text{green}}$} \\
\textbf{Jitter}         & \textbf{0.0347} & \textbf{$\mu_{\text{red}} > \mu_{\text{blue}}$}   & 0.0667 & $\mu_{\text{green}} > \mu_{\text{blue}}$ \\
\textbf{Max Intensity}           & 0.1260 & $\mu_{\text{red}} > \mu_{\text{blue}}$ & \textbf{0.0023} & \textbf{$\mu_{\text{green}} > \mu_{\text{blue}}$} \\
\bottomrule
\end{tabular}}
\caption{T-test results comparing red vs.\ blue and green vs.\ blue examples for audio features with $\alpha = 0.05$. Statistically significant results are bolded. See Appendix~\ref{appendix:full-feature-comparison} for full table.}
\label{tab:ttest-combined}
\vspace{-15pt}
\end{table}

%% file: tables/au-ttests-small.tex
\begin{table}[t]
\resizebox{\columnwidth}{!}{
\setlength{\tabcolsep}{6pt}
\centering
\begin{tabular}{llcc|cc}
\toprule
\textbf{AU} & \textbf{p (R vs B)} & \textbf{Dir} & \textbf{p (G vs B)} & \textbf{Direction} \\
\midrule
\textbf{AU04} & \textbf{0.0106} & \textbf{red $>$ blue}   & 0.3682 & green $>$ blue \\
\textbf{AU12} & \textbf{0.0174} & \textbf{blue $>$ red}   & 0.8977 & green $>$ blue \\
\textbf{AU05} & 0.1837 & blue $>$ red   & \textbf{<0.0001} & \textbf{blue $>$ green} \\
\bottomrule
\end{tabular}}
\caption{T-test results comparing AU activation rates between red vs.\ blue and green vs.\ blue  with $\alpha = 0.05$. Statistically significant results are bolded. See Appendix~\ref{appendix:full-feature-comparison} for full table}
\label{tab:au-ttests-small}
\vspace{-15pt}
\end{table}

%% file: tables/umap.tex

\begin{figure}
\centering
\includegraphics[width=1.0\columnwidth]{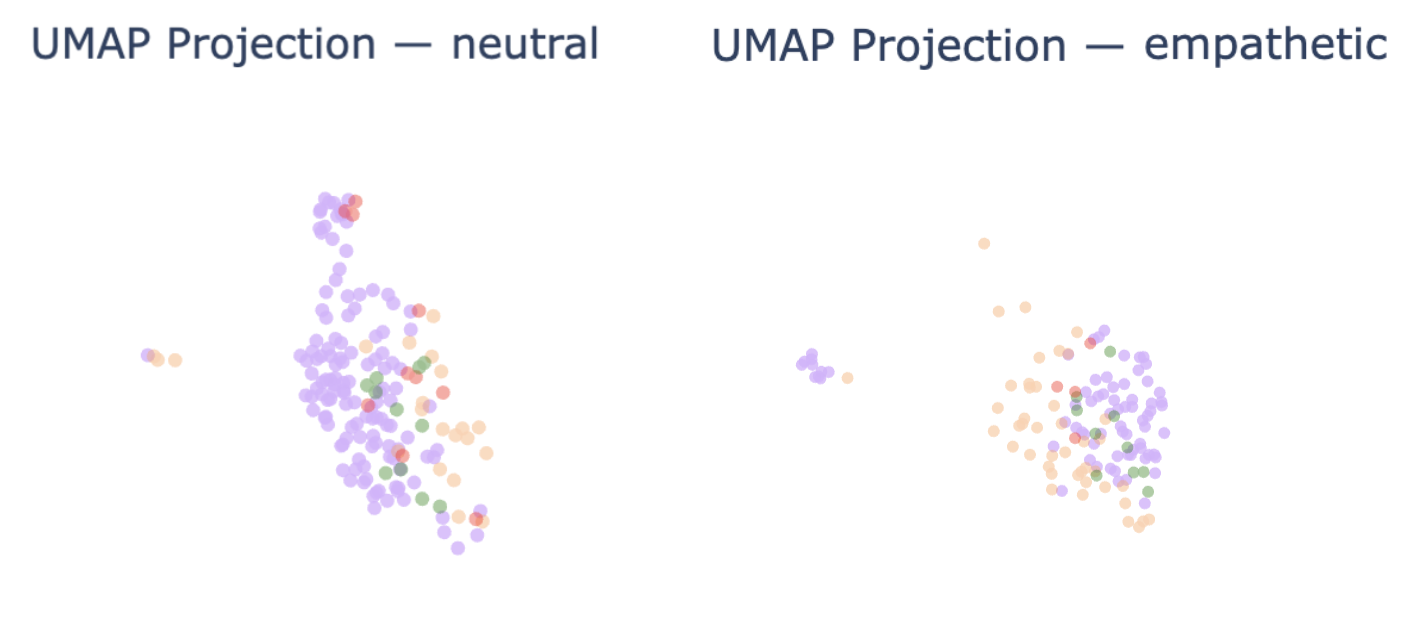}
\caption{UMAP of text-only embeddings for empathetic (left) vs. neutral (right) examples, colored by modality disagreement; red and green points cluster near the decision boundary, marking ambiguous cases.}
\label{fig:umap}
\vspace{-0.7em}
\end{figure}

%% file: tables/entropy.tex
\begin{table}
\centering
\small 
\begin{tabular}{lcc}
\toprule
\textbf{Quadrant} & \textbf{Mean Entropy} & \textbf{St. Dev} \\
\midrule
\colorbox{lightred}{\textcolor{darkred}{Red}}     & 0.670 & 0.017\\
\colorbox{lightblue}{\textcolor{darkblue}{Blue}}   & 0.259 & 0.222 \\
\colorbox{lightyellow}{\textcolor{darkgold}{Yellow}} & 0.347 & 0.196 \\
\colorbox{lightgreen}{\textcolor{darkgreen}{Green}}  & 0.565 & 0.192\\
\bottomrule
\end{tabular}
\caption{Mean entropy of the fusion model grouped by quadrant }
\label{tab:entropy}
\vspace{-15pt}
\end{table}

%% file: sections/04-experiment-2.tex
We further assess whether model disagreements stem from data ambiguity using a human annotation study that tests whether examples from the two disagreement regions (red and green quadrants) are equally challenging for annotators.

\paragraph{Annotation Setup.} We sample 204 examples evenly split across the four quadrants of each Figure~\ref{fig:conf} modality plot. 
For each example, annotators provide a binary judgment (empathetic or neutral) from a unimodal signal, then a judgment from the full multimodal version (instructions in Appendix~\ref{appendix:annotation}), allowing us to track how human predictions shift with additional modality signals and to understand the cognitive burden of multimodal integration.
All examples were annotated by one author and one external annotator. Table~\ref{tab:quadrant_examples} in the Appendix
showcases frames and transcripts for four examples, along with annotator judgments. 

\input{tables/kappa}

\paragraph{Results.}
Annotator \textit{disagreement}, measured with Cohen's Kappa~\cite{cohen1960coefficient}, can signal complex phenomena in examples~\cite{jiang-marneffe-2022-investigating, pavlick} such as uncertainty in meaning leading to discrepancies in reasoning.
In disagreement regions (red and green), we see a \textit{decrease} in annotator agreement between unimodal and multimodal judgments (Table~\ref{tab:kappa-by-color}), indicating that humans diverge when weighing signals across modalities. 
In contrast, annotator agreement \textit{improves} upon examples where unimodal and multimodal model predictions are in agreement, supporting our hypothesis that these examples are relatively unambiguous and can be reliably interpreted once the full context is available (Table~\ref{tab:kappa-by-color}). 

We repeat the analysis on the subset of clips containing at least six tokens (the median utterance length in \empspeech). Examples below the median often include short phrases and backchannels, while examples above the median are often complete sentences with richer lexical and syntactic structure. As shown in Table~\ref{tab:kappa-long}, the red and green quadrant utterances continue to exhibit a substantial drop in Cohen's $\kappa$ compared to blue and yellow quadrant utterances, which exhibit substantial gains with additional information.  
These results collectively corroborate our hypothesis that modality disagreement can serve as a valuable signal for identifying ambiguous, challenging, or complex instances that are also difficult for human annotators.
\input{tables/annot_long}

%% file: tables/kappa.tex
\begin{table}[t]
\centering
\resizebox{\columnwidth}{!}{
\begin{tabular}{lccc}
\toprule
\textbf{Quadrant} & \textbf{Unimodal Judgment} & \textbf{Multimodal Judgment} &\textbf{$\Delta$}\\
\midrule
\colorbox{lightred}{\textcolor{darkred}{Red}}    & 0.301 & 0.164 & -0.137\\
\colorbox{lightblue}{\textcolor{darkblue}{Blue}}   & 0.379 & 0.646 &0.267 \\
\colorbox{lightyellow}{\textcolor{darkgold}{Yellow}} & 0.225 & 0.329 & 0.104 \\
\colorbox{lightgreen}{\textcolor{darkgreen}{Green}}  & 0.482 & 0.218 & -0.264\\
\bottomrule
\end{tabular}}
\caption{Cohen’s Kappa between internal and external annotators, computed separately for each quadrant and prediction round.}
\label{tab:kappa-by-color}
\vspace{-15pt}
\end{table}

%% file: tables/annot_long.tex
\begin{table}[t]
\centering
\resizebox{\columnwidth}{!}{
\begin{tabular}{lccc}
\toprule
\textbf{Quadrant} & \textbf{Unimodal Judgment} & \textbf{Multimodal Judgment} &\textbf{$\Delta$}\\
\midrule
\colorbox{lightred}{\textcolor{darkred}{Red}}    & 0.347 & 0.143 & -0.204\\
\colorbox{lightblue}{\textcolor{darkblue}{Blue}}   & 0.364 & 0.533 &0.169 \\
\colorbox{lightyellow}{\textcolor{darkgold}{Yellow}} & 0.304 & 0.548 & 0.244 \\
\colorbox{lightgreen}{\textcolor{darkgreen}{Green}}  & 0.573 & 0.329 & -0.244\\
\bottomrule
\end{tabular}}
\caption{Cohen’s Kappa between internal and external annotators for examples of at least six tokens (the dataset median), computed separately for each quadrant
and prediction round.}
\label{tab:kappa-long}
\vspace{-15pt}
\end{table}

%% file: sections/05-conclusion.tex
We have demonstrated how disagreement, both between modalities and between humans and models, can serve as a diagnostic lens to understand the complexity of multimodal empathy detection, challenging the assumption that more signals from other modalities reliably yields better performance.
Our analysis reveals that disagreement between unimodal and multimodal models is often not arbitrary, but instead marks the presence of subtle, ambiguous, or context-sensitive cues that challenge fusion models and human annotators alike.

While our study focuses on speaker-centric empathy (evaluating speakers’ empathic expression), our diagnostic can be generalized to listener-centric tasks, which dominate existing empathy datasets and capture listeners’ emotional responses to each utterance (Appendix \ref{appendix:data-comp}).
These findings emphasize the necessity for high-quality annotation in socially complex tasks like empathy detection, where model errors may reflect genuine human uncertainty or disagreement. This framework provides a scalable method for identifying ambiguity and enhancing model reliability, especially in recognizing complex emotional states.

Beyond diagnosis, disagreement offers a foundation for improving multimodal learning. Cross-modal conflict can guide labeling efforts toward informative and ambiguous examples, making annotation more efficient when resources are limited. Patterns of disagreement can also inform curriculum design~\cite{qian}, where models first learn from consistent, low-disagreement examples before tackling more ambiguous ones to build nuanced reasoning and robustness. Furthermore, insights from disagreement can inspire more adaptive fusion approaches that dynamically re-weight or downplay misleading modalities when they conflict~\cite{huang}, reducing over-reliance on a single signal. High-disagreement examples can also serve as realistic adversarial test cases that expose systematic vulnerabilities and strengthen fusion strategies under genuine multimodal conflict~\cite{yang2022}.
Finally, this diagnostic perspective extends beyond empathy detection to other socially complex tasks such as persuasion~\cite{bai}, rapport~\cite{baihaqi}, or sarcasm~\cite{Zhou_2024}, where multimodal cues and subjective judgments often diverge. In such settings, disagreement between unimodal and fusion models highlights genuinely ambiguous cases that can guide targeted annotation, evaluation, and model refinement.

Ultimately, treating disagreement as a meaningful signal rather than an error reframes how we evaluate and improve multimodal models. By revealing when and why models diverge, this perspective lays the foundation for building systems that reason more like humans do. Beyond empathy detection, this framework also opens broader pathways toward socially intelligent multimodal systems that can recognize uncertainty, resolve conflicting evidence, and adapt their reasoning to the inherent ambiguity of human affective communication.
%

%% file: sections/06-limitations.tex
We acknowledge several limitations in our study. Our analyses are based on a limited dataset and a small number of human annotators. Given that empathy is inherently subjective, annotations may vary due to individual interpretations, potentially introducing biases rather than reflecting universal properties of the data. Additionally, we rely upon a single dataset, and future work should investigate whether the patterns we observe hold across other datasets and domains.

Our data is also derived from U.S.-based, English-language television and interview content. As such, the generalizability of our findings to multilingual or culturally diverse settings may be limited. Future research should investigate these patterns in varied cultural and linguistic environments to better assess the broader applicability of our conclusions.

%% file: sections/07-ethicalconsiderations.tex
We used a publicly available dataset and strictly use open-source models for analysis. 

All annotations were conducted by an author and an individual affiliated with the research team. No participants were recruited via crowdsourcing or external platforms, and no monetary compensation was provided, as the annotators were contributing in a research capacity. We provided detailed information on what we asked the annotators to annotate and how we planned to use the data. The annotators willingly agreed to participate with full knowledge of the task. No sensitive or identifying information was collected from annotators.

We note that empathy expression may vary across cultures, and our findings may not generalize to non-English or non-Western contexts. We encourage future work to explore these questions in more diverse settings.

We will release all code and experimental resources at \url{https://github.com/mayasrikm/multimodal-empathy-disagreement}
to support reproducibility.

%% file: sections/08-ack.tex
This work was supported in part by a grant from the Columbia Center of AI Technology (CAIT) in collaboration with Amazon. 
The views, opinions and/or findings expressed are solely those of the authors.

%% file: sections/appendix.tex
\section{Dataset}
\label{appendix:data}
\subsection{Dataset Specifications}
We use a multimodal empathy dataset \cite{chen} consisting of 346 English-language videos totaling approximately 53 hours, collected from YouTube between 2020 and 2022 using keywords like “empathy” and “empathetic training.” The dataset includes empathy training sessions, therapy roleplays, interviews, TED Talks, and TV/movie scenes, comprising both acted (62\%) and spontaneous (38\%) speech. Each video was labeled by at least three expert annotators as either empathetic or neutral, with final labels determined by majority vote. Metadata such as speaker gender, topic, and emotional context was manually annotated, covering themes like therapy, parenting, workplace dynamics, and social relationships. From this collection, a subset of 65 videos was transcribed, diarized, and manually re-aligned using Praat to ensure accurate speaker segmentation and time alignment. This process resulted in 1,718 annotated segments with speaker labels, timestamps, transcripts, and empathy stage annotations, enabling fine-grained analysis of empathy in naturalistic and semi-scripted settings. The median utterance length of the dataset is six tokens. Table~\ref{tab:token_examples} shows example utterances below and above the median; examples below the median often include short phrases and backchannels, while examples above the median are often complete sentences with richer lexical and syntactic structure.
\input{tables/utterances.tex}
\subsection{Dataset Comparison}
\label{appendix:data-comp}
To the best of our knowledge, our work is the first to evaluate multimodal disagreement on speaker-centric empathy detection datasets. 
Most publicly available empathy datasets (such as \abr{EmpathicStories++} \cite{shen} and \abr{OMG-Empathy} \cite{barros}) are fundamentally structured around listener response, not speaker expression. In these datasets, the task is to predict how empathetic a listener feels after hearing a story, rather than to assess whether the speaker themselves is expressing empathy. 
For instance, in \abr{EmpathicStories++}, participants record personal stories and then rate their own emotional responses, framing empathy as a \textit{reaction} to the content rather than as a property of the speaker’s delivery. 
Similarly, \abr{OMG-Empathy} evaluates listener self-reported affective states following brief monologues, again focusing on perceived empathy rather than expressed empathy. This distinction matters because listener-focused tasks inherently entangle speaker behavior with listener subjectivity, making it difficult to isolate which cues (textual, audial, or visual) are directly responsible for empathy expression. In contrast, the dataset we use in this study~\cite{chen} is one of the only accessible resources that explicitly asks annotators to evaluate the \textit{speaker’s} empathy, based solely on the speech segment itself, across modalities. This framing allows us to analyze how empathy is expressed in real time by the speaker, independent of listener interpretation, and enables direct comparisons between modalities on their ability to convey empathetic intent. 
Our current dataset offers a uniquely valuable lens into the structure of empathy as a speaker-side communicative behavior: something that remains underexplored in the literature. 
In future work, our modality-disagreement diagnostic could be used to flag nuanced, high-ambiguity segments that challenge \textit{listener} empathy models. 
They could serve as an effective proxy for identifying segments that elicit high listener variance in empathy judgments, enabling targeted annotation and model refinement on exactly those ambiguous utterances where listener-centric prediction systems struggle most.

\section{Model Training Details}
Data was split into train, test and validation sets using random sampling, with an 80-10-10 split. We run fine-tuning and inference for all open-source models on an A100 GPU in Google Colab. 
\subsection{Unimodal Model Training Details}
\label{appendix:uni-model-details}
Each model is trained on a binary empathy classification task using precomputed 768-dimensional embeddings. 
We freeze all but the final two transformer layers and train for fifteen epochs with a learning rate of 5e-6 and batch size of eight. 
\subsection{Fusion Model Details}
\label{appendix:fusion-model-details}
Each unimodal model representation is independently gated and passed through an additive attention mechanism that computes modality-specific weights. The weighted embeddings are aggregated and classified using a three-layer feedforward network with max pooling. The fusion model is trained for ten epochs using a learning rate of 1e-4 and includes modality dropout during training. 
To characterize how the model balances each of the three modalities at inference time, we computed the per-modality gate-weight distributions over the full test set (Table ~\ref{tab:fusion-details}). The mean gate weights indicate that our model allocates substantial importance to each modality, with only a slight preference toward audio and video. The high variance also shows that the model dynamically adapts its reliance on each modality on a per‐sample basis. Thus, our fusion model draws substantially on all three streams; no single modality is systematically favored. 

\input{tables/fusion-details.tex}

\section{Annotation Instructions} \label{appendix:annotation}
We employed two annotators, one of the paper's authors and an non-author, both fluent English speakers based in the United States. No additional demographic information was collected, as the annotation was conducted internally for research purposes.

Annotators were asked to provide two judgments per example, labeling each as either empathetic or neutral (Figure \ref{fig:annotation}). A excerpt describing empathy (drawn from the Encyclopedia of Social Psychology, Volume 1, \cite{baumeister}) was provided to ensure a consistent conceptual foundation for annotation: 
\begin{quote}
\small
Empathy is often defined as understanding another person’s experience by imagining oneself in that other person’s situation: One understands the other person’s experience as if it were being experienced by the self, but without the self actually experiencing it. There are three commonly studied components of emotional empathy. The first is feeling the same emotion as another person (sometimes attributed to emotional contagion, e.g., unconsciously “catching” someone else’s tears and feeling sad oneself). The second component, personal distress, refers to one’s own feelings of distress in response to perceiving another’s plight. The third emotional component, feeling compassion for another person, is the one most frequently associated with the study of empathy. Cognitive empathy refers to the extent to which we perceive or have evidence that we have successfully guessed someone else’s thoughts and feelings.
\end{quote}

Annotators were given an annotation flag indicating which modality to use for the first pass; for instance, if the flag was text, only the transcript was to be used to make the first prediction. After submitting the first judgment, annotators were then given access to the full video, including all available audio, visual, and textual information. They were then asked to provide a second prediction.
\input{tables/annot}
\section{Full Feature Comparisons} \label{appendix:full-feature-comparison}
Tables \ref{tab:audio-feature-significance}, \ref{tab:ttest-combined-full} and \ref{tab:au-ttests-red-green} provide additional results from the t-tests comparing examples across different confidence quadrants.  Table \ref{tab:audio-feature-significance} provides an internal comparison between the disagreement quadrants. 
Table~\ref{tab:ttest-combined-full} presents the full version of the audio feature comparisons summarized in Table~\ref{tab:ttest-combined}.
Table~\ref{tab:au-ttests-red-green} expands on the facial feature comparisons shown in Table~\ref{tab:au-ttests-small}.
\input{tables/red-green}

\input{tables/audio-ttests}
\input{tables/au-ttests}

\section{Feature Distributions}
Figures~\ref{fig:audio-plot} and~\ref{fig:au-activation} visualize the distributions of key features across confidence quadrants. Figure~\ref{fig:audio-plot} presents the distribution of selected audio features (e.g., pitch, intensity) for red, green, and blue examples, highlighting acoustic patterns associated with model disagreement. 
Figure~\ref{fig:au-activation} shows activation rates for facial Action Units (AUs) in red, green, and blue examples, illustrating how specific facial expressions vary across agreement conditions. These visualizations complement the statistical comparisons reported in Tables \ref{tab:ttest-combined-full} and \ref{tab:au-ttests-red-green}, providing a more interpretable view of the underlying feature dynamics.

\input{tables/distrib}

\section{Quadrant Examples}
Table \ref{tab:quadrant_examples} shows video frames, transcripts, annotator judgments, and the true labels for examples from each confidence plot quadrant. These examples illustrate that disagreement quadrants often contain more ambiguous instances for both humans and models where cues from different modalities may conflict, while examples from agreement quadrants typically display alignment between modalities.

\input{tables/frame.tex}

%% file: tables/utterances.tex
\begin{table*}[ht]
  \centering
  \begin{tabular}{p{0.48\textwidth} p{0.48\textwidth}}
    \hline
    \textbf{< 6 Tokens} & \textbf{\boldmath$\ge6$ Tokens}\unboldmath\\
    \hline
    \textit{“there’s no way”} (3 tokens)
      & \textit{“No, no he’s a good guy go easy on him he’s lost his son, Fabio”} (15 tokens)\\
    \textit{“You lost it?”} (3 tokens)
      & \textit{“You kids have the biggest hearts I’ve ever seen.”} (9 tokens)\\
    \textit{“I can understand that.”} (4 tokens)
      & \textit{“congrats my dude, on everything man”} (6 tokens)\\
    \hline
  \end{tabular}
  \caption{Example utterances with fewer than six tokens (left) versus at least six tokens (right).}
  \label{tab:token_examples}
\end{table*}

%% file: tables/fusion-details.tex
\begin{table}[ht]
  \centering
  \begin{tabular}{lrr}
    \toprule
    \textbf{Modality} & \textbf{Mean} & \textbf{Standard Deviation} \\
    \midrule
    \textbf{Text}     & 0.430 & 0.297              \\
    \textbf{Audio}    & 0.502 & 0.227              \\
    \textbf{Video}    & 0.476 & 0.315              \\
    \bottomrule
  \end{tabular}
  \caption{Per-modality gate-weight distributions over the full test set.}
  \label{tab:fusion-details}
\end{table}

%% file: tables/annot.tex
\begin{figure*}[t]
\centering
\includegraphics[width=0.8\textwidth]{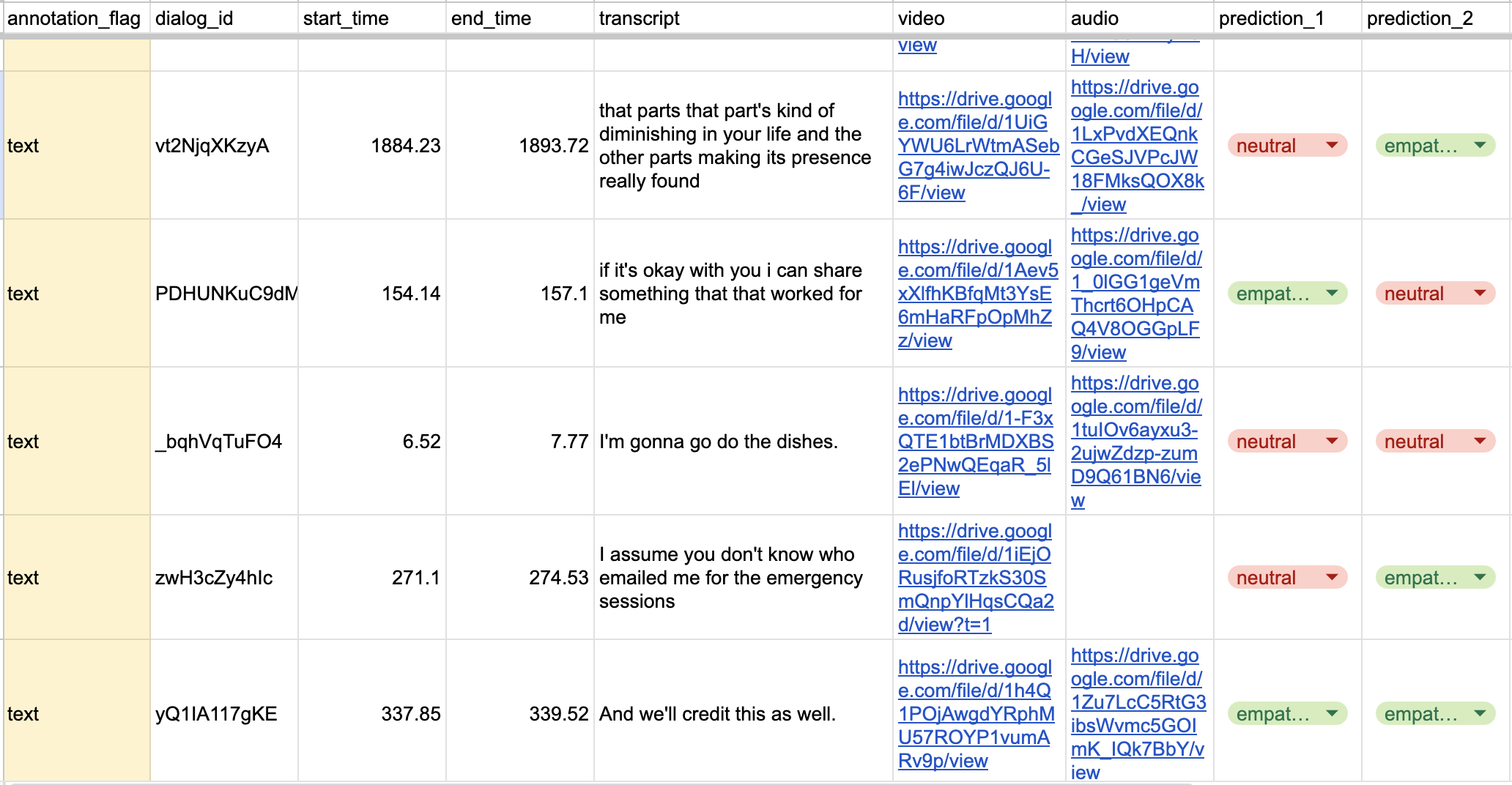}
\caption{Annotation interface}
\label{fig:annotation}
\end{figure*}

%% file: tables/red-green.tex
\begin{table*}[ht]
\centering
\scriptsize
\resizebox{0.68\textwidth}{!}{
\begin{tabular}{|l|c|c|c|}
\hline
\textbf{Feature} & \textbf{t-stat} & \textbf{p-value} & \textbf{Mean Comparison} \\
\hline
\textbf{Mean Pitch}      &  
\textbf{2.453} & \textbf{0.0159}  & \boldmath$\mu_{\text{red}} > \mu_{\text{green}}$ \\
\textbf{Max Intensity}   & \textbf{-2.124} & \textbf{0.0366} & \boldmath$\mu_{\text{green}} > \mu_{\text{red}}$ \\
\textbf{Max Pitch}       &  \textbf{2.016} & \textbf{0.046}5 & \boldmath$\mu_{\text{red}} > \mu_{\text{green}}$ \\
\textbf{Min Pitch}       &  \textbf{2.007} & \textbf{0.0475}  & \boldmath$\mu_{\text{red}} > \mu_{\text{green}}$ \\
valence         & -1.908 & 0.0593 & $\mu_{\text{green}} > \mu_{\text{red}}$ \\
arousal         &  1.827 & 0.0705 & $\mu_{\text{red}} > \mu_{\text{green}}$ \\
speaking\_rate  &  1.773 & 0.0807  & $\mu_{\text{red}} > \mu_{\text{green}}$ \\
dominance       &  1.712 & 0.0899  & $\mu_{\text{red}} > \mu_{\text{green}}$ \\
Shimmer         &  0.773 & 0.4416  & $\mu_{\text{red}} > \mu_{\text{green}}$ \\
Jitter          &  0.622 & 0.5355  & $\mu_{\text{red}} > \mu_{\text{green}}$ \\
Mean Intensity  &  0.544 & 0.5886 & $\mu_{\text{red}} > \mu_{\text{green}}$ \\
HNR             &  0.508 & 0.6129  & $\mu_{\text{red}} > \mu_{\text{green}}$ \\
Min Intensity   & -0.429 & 0.6685  & $\mu_{\text{green}} > \mu_{\text{red}}$ \\
\hline
\end{tabular}
}
\captionof{table}{T-test results comparing audio features between red and green examples. Statistically significant results are bolded. }
\label{tab:audio-feature-significance}
\end{table*}

%% file: tables/audio-ttests.tex
\begin{table*}[ht]
\centering
\scriptsize
\resizebox{0.9\textwidth}{!}{%
\begin{tabular}{|l|c|c|c|c|}
\hline
\textbf{Feature} & \textbf{p (Red vs Blue)} & \textbf{Direction} & \textbf{p (Green vs Blue)} & \textbf{Direction} \\
\hline
\textbf{valence}        & \textbf{0.0047} & \boldmath$\mu_{\text{blue}} > \mu_{\text{red}}$ & 0.5166 & $\mu_{\text{green}} > \mu_{\text{blue}}$ \\
\textbf{arousal}        & \textbf{0.0065} & \boldmath$\mu_{\text{blue}} > \mu_{\text{red}}$ & \textbf{0.0136} & \boldmath$\mu_{\text{blue}} > \mu_{\text{green}}$ \\
\textbf{Mean Pitch}     & \textbf{0.0100} & \boldmath$\mu_{\text{blue}} > \mu_{\text{red}}$ & \textbf{0.0001} & \boldmath$\mu_{\text{blue}} > \mu_{\text{green}}$ \\
\textbf{dominance}      & \textbf{0.0108} & \boldmath$\mu_{\text{blue}} > \mu_{\text{red}}$ & 0.0667 & $\mu_{\text{blue}} > \mu_{\text{green}}$ \\
\textbf{Min Pitch}      & \textbf{0.0333} & \boldmath$\mu_{\text{blue}} > \mu_{\text{red}}$ & \textbf{0.0001} & \boldmath$\mu_{\text{blue}} > \mu_{\text{green}}$ \\
\textbf{Jitter}         & \textbf{0.0347} & \boldmath$\mu_{\text{red}} > \mu_{\text{blue}}$ & 0.0667 & $\mu_{\text{green}} > \mu_{\text{blue}}$ \\
Max Intensity           & 0.1260 & $\mu_{\text{red}} > \mu_{\text{blue}}$ & \textbf{0.0023} & \boldmath$\mu_{\text{green}} > \mu_{\text{blue}}$ \\
Mean Intensity          & 0.1599 & $\mu_{\text{red}} > \mu_{\text{blue}}$ & 0.5329 & $\mu_{\text{blue}} > \mu_{\text{green}}$ \\
HNR                     & 0.2217 & $\mu_{\text{blue}} > \mu_{\text{red}}$ & 0.2055 & $\mu_{\text{blue}} > \mu_{\text{green}}$ \\
speaking\_rate          & 0.2723 & $\mu_{\text{blue}} > \mu_{\text{red}}$ & 0.9991 & $\mu_{\text{green}} > \mu_{\text{blue}}$ \\
Shimmer                 & 0.4122 & $\mu_{\text{red}} > \mu_{\text{blue}}$ & 0.1541 & $\mu_{\text{blue}} > \mu_{\text{green}}$ \\
Max Pitch               & 0.6845 & $\mu_{\text{red}} > \mu_{\text{blue}}$ & 0.2647 & $\mu_{\text{blue}} > \mu_{\text{green}}$ \\
Min Intensity           & 0.7999 & $\mu_{\text{blue}} > \mu_{\text{red}}$ & 0.1571 & $\mu_{\text{blue}} > \mu_{\text{green}}$ \\
\hline
\end{tabular}
}
\captionof{table}{T-test results comparing audio features between red vs.\ blue and green vs.\ blue examples. Statistically significant p-values are bolded.}
\label{tab:ttest-combined-full}
\end{table*}

%% file: tables/au-ttests.tex
\begin{table*}[t]
\centering
\scriptsize
\resizebox{\textwidth}{!}{
\begin{tabular}{|l|c|c|c|c|}
\hline
\textbf{AU} & \textbf{p (Red vs Blue)} & \textbf{Direction} & \textbf{p (Green vs Blue)} & \textbf{Direction} \\
\hline
\textbf{AU04: Brow Lowerer}         & \textbf{0.0106} & \textbf{red $>$ blue}   & 0.3682 & green $>$ blue \\
\textbf{AU12: Lip Corner Puller}    & \textbf{0.0174} & \textbf{blue $>$ red}   & 0.8977 & green $>$ blue \\
\textbf{AU05: Upper Lid Raiser}     & 0.1837 & blue $>$ red   & \textbf{<0.0001} & \textbf{blue $>$ green} \\
AU17: Chin Raiser           & 0.2256 & red $>$ blue   & 0.9802 & blue $>$ green \\
AU10: Upper Lip Raiser      & 0.2275 & blue $>$ red   & 0.6700 & green $>$ blue \\
AU45: Blink                 & 0.3200 & blue $>$ red   & 0.7462 & green $>$ blue \\
AU07: Lid Tightener         & 0.3252 & blue $>$ red   & 0.9318 & blue $>$ green \\
AU14: Dimpler               & 0.4593 & red $>$ blue   & 0.0652 & green $>$ blue \\
AU20: Lip Stretcher         & 0.5701 & blue $>$ red   & 0.7907 & blue $>$ green \\
AU09: Nose Wrinkler         & 0.6211 & blue $>$ red   & 0.7639 & green $>$ blue \\
AU25: Lips Part             & 0.6227 & blue $>$ red   & 0.7492 & blue $>$ green \\
AU01: Inner Brow Raiser     & 0.6529 & blue $>$ red   & 0.4674 & green $>$ blue \\
AU23: Lip Tightener         & 0.6630 & red $>$ blue   & 0.3474 & green $>$ blue \\
AU28: Lip Suck              & 0.6735 & red $>$ blue   & 0.9846 & green $>$ blue \\
AU26: Jaw Drop              & 0.6851 & red $>$ blue   & 0.4596 & blue $>$ green \\
AU06: Cheek Raiser          & 0.7097 & blue $>$ red   & 0.3201 & green $>$ blue \\
AU15: Lip Corner Depressor  & 0.9528 & red $>$ blue   & 0.4834 & green $>$ blue \\
AU02: Outer Brow Raiser     & 0.9647 & blue $>$ red   & 0.6677 & green $>$ blue \\
\hline
\end{tabular}
}
\captionof{table}{T-test results comparing AU activation rates between red vs.\ blue and green vs.\ blue. Bolded p-values are statistically significant.}
\label{tab:au-ttests-red-green}
\end{table*}

%% file: tables/distrib.tex
\begin{figure*}[t]
\centering
\includegraphics[width=\textwidth]{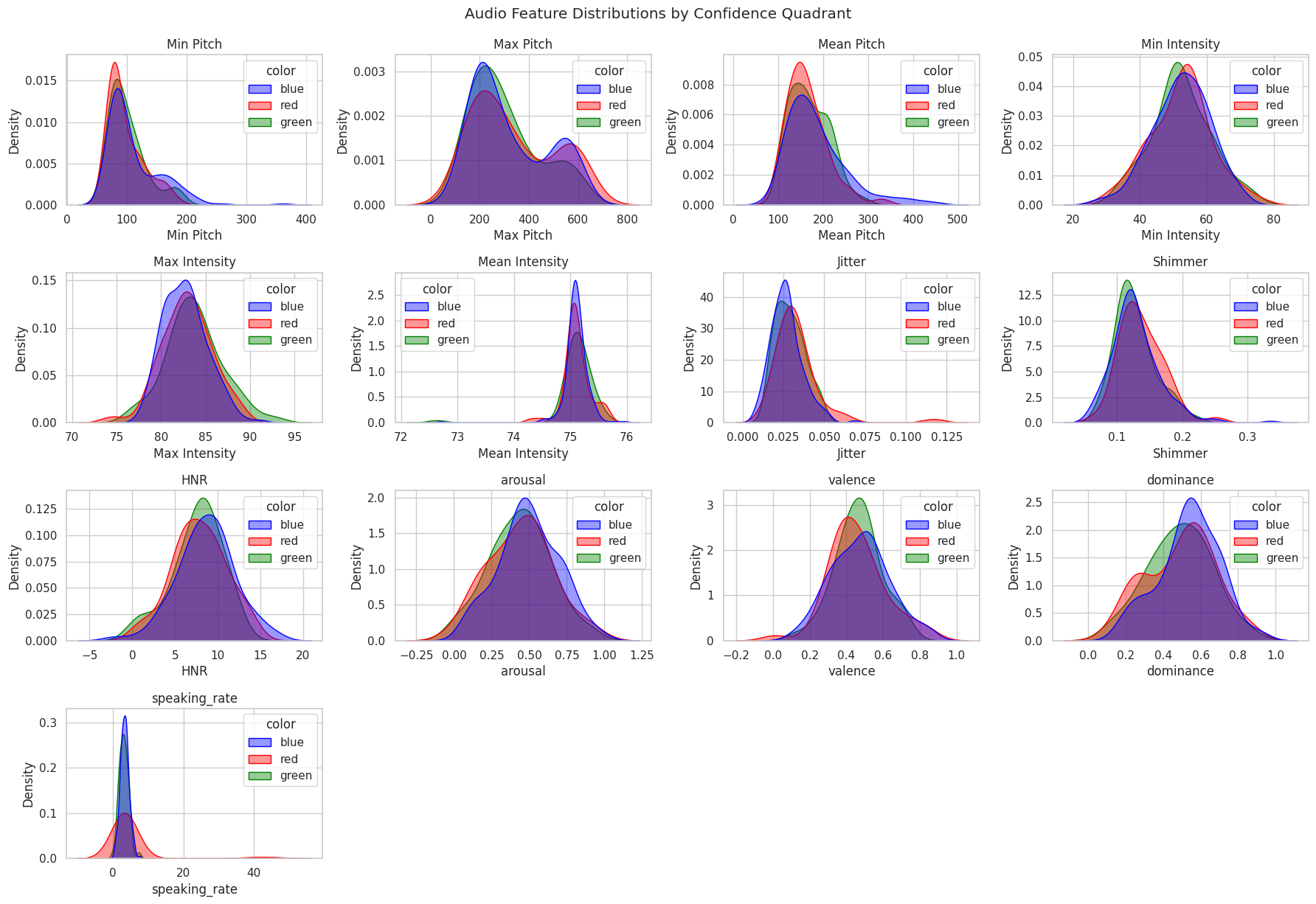}
\caption{Distribution of audio features for red, green and blue examples across the confidence quadrants. Red examples are those correctly classified by the unimodal audio model but misclassified by the multimodal model; green examples represent the reverse. Blue examples represent those correctly classified by both the unimodal audio model and the multimodal model. Significant differences appear in pitch and intensity-based features.}
\label{fig:audio-plot}
\end{figure*}

\begin{figure*}[t]
\centering
\includegraphics[width=\textwidth]{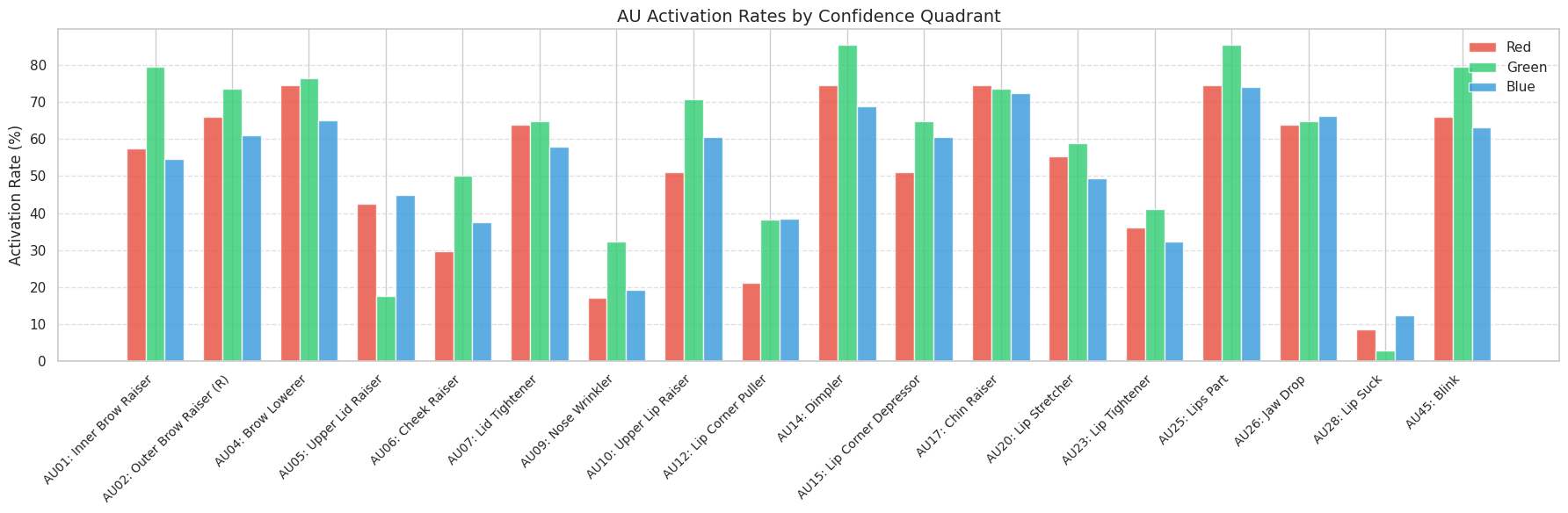}
\caption{AU activation rates for red, green, and blue examples. Red bars indicate examples where the unimodal visual model predicted correctly but the multimodal model did not (Red: Unimodal $>$ 0.5, Multimodal $<$ 0.5). Green bars show the reverse. Blue bars indicate examples where both the unimodal and multimodal models correctly predicted the label.}
\label{fig:au-activation}
\end{figure*}

%% file: tables/frame.tex
\renewcommand{\arraystretch}{3} 
\begin{table*}[ht]
  \centering
  \setlength{\tabcolsep}{12pt} 
  \begin{tabular}{|p{0.48\textwidth}|p{0.48\textwidth}|}
    \hline
    \begin{minipage}[t]{\linewidth}
      \centering
      \vspace{1pt} 
      \includegraphics[width=0.63\linewidth]{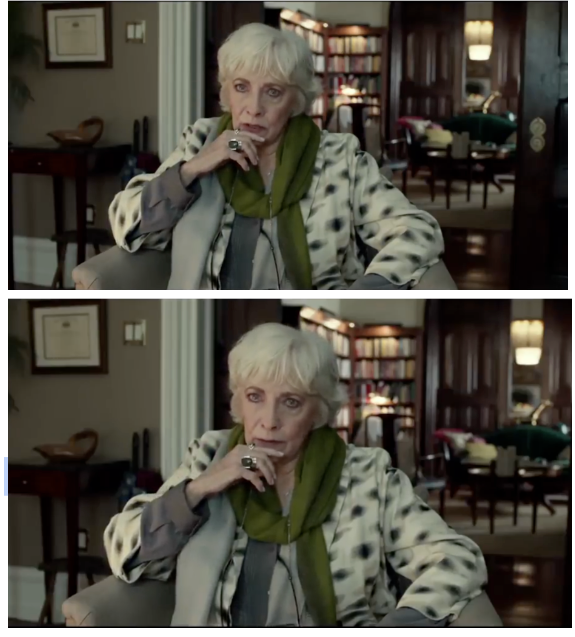}\\[2ex]
      \textbf{Transcript:} "I assume you don't know who emailed me for the emergency sessions"\\[1ex]
      \textbf{Quadrant:} \colorbox{lightred}{\textcolor{darkred}{Red}}\\[0.5ex]
      \textbf{True Label:} Empathetic\\[0.5ex]
      \textbf{Annotator 1:} Neutral\\[0.5ex]
      \textbf{Annotator 2:} Empathetic\\[5pt]
      \vspace{2pt} 
    \end{minipage}
    &
    \begin{minipage}[t]{\linewidth}
      \centering
      \vspace{1pt} 
      \includegraphics[width=0.63\linewidth]{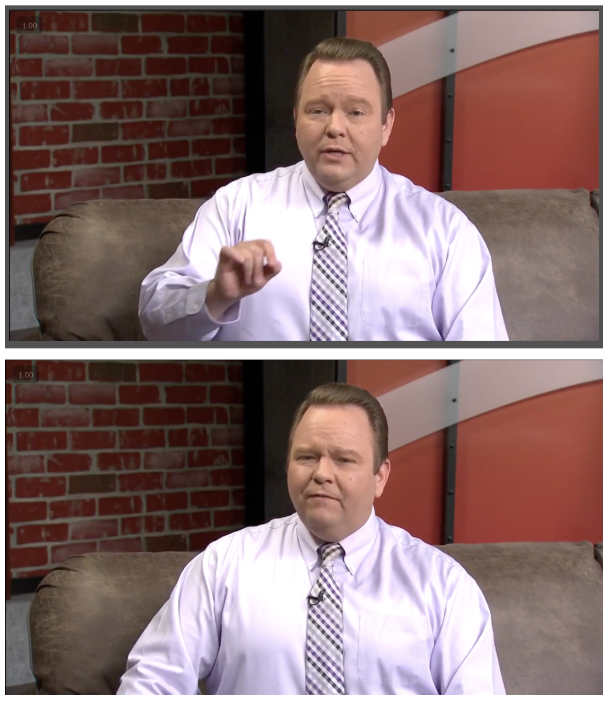}\\[2ex]
      \textbf{Transcript:} "In fact research suggests we spend about 55 percent of our day..."\\[1ex]
      \textbf{Quadrant:} \colorbox{lightblue}{\textcolor{darkblue}{Blue}}\\[0.5ex]
      \textbf{True Label:} Neutral\\[0.5ex]
      \textbf{Annotator 1:} Neutral\\[0.5ex]
      \textbf{Annotator 2:} Neutral\\[5pt]
      \vspace{2pt} 
    \end{minipage}
    \\ \hline
    \begin{minipage}[t]{\linewidth}
      \centering
      \vspace{1pt} 
      \includegraphics[width=0.63\linewidth]{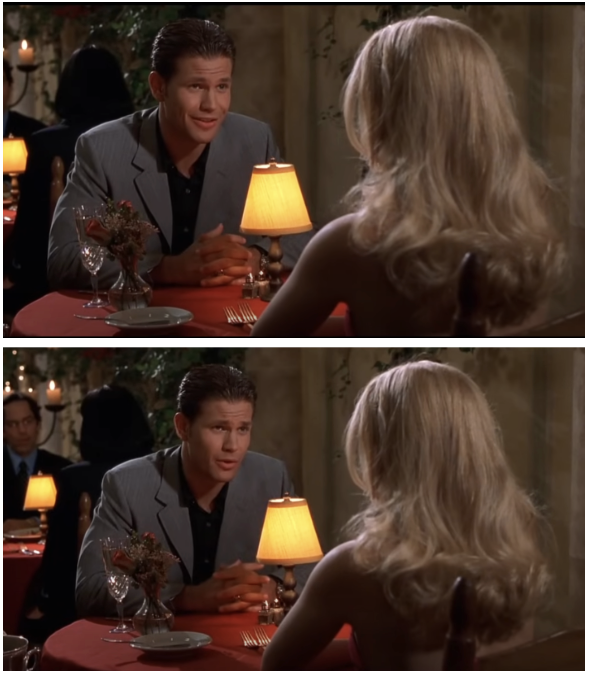}\\[2ex]
      \textbf{Transcript:} "One of the reasons I wanted to come here tonight was to discuss our future."\\[1ex]
      \textbf{Quadrant:} \colorbox{lightyellow}{\textcolor{darkgold}{Yellow}}\\[0.5ex]
      \textbf{True Label:} Neutral\\[0.5ex]
      \textbf{Annotator 1:} Empathetic\\[0.5ex]
      \textbf{Annotator 2:} Empathetic\\[5pt]
      \vspace{2pt} 
    \end{minipage}
    &
    \begin{minipage}[t]{\linewidth}
      \centering
      \vspace{1pt} 
      \includegraphics[width=0.63\linewidth]{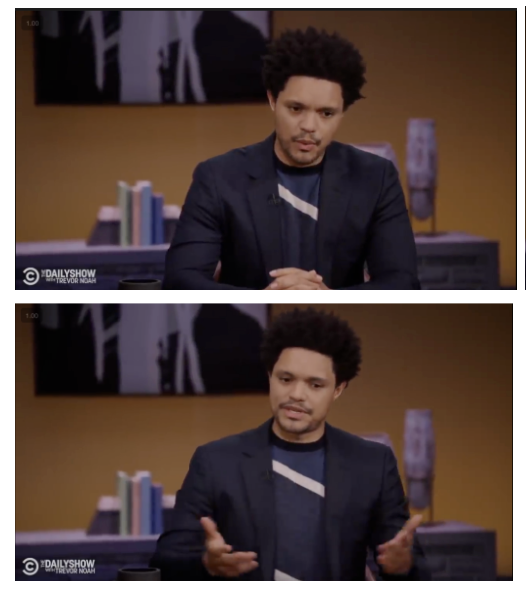}\\[2ex]
      \textbf{Transcript:} "It's good to have you here um especially to talk about a topic that i think is one of the more sensitive topics that we we're discussing in society today..."\\[1ex]
      \textbf{Quadrant:} \colorbox{lightgreen}{\textcolor{darkgreen}{Green}}\\[0.5ex]
      \textbf{True Label:} Empathetic\\[0.5ex]
      \textbf{Annotator 1:} Neutral\\[0.5ex]
      \textbf{Annotator 2:} Empathetic\\[5pt]
      \vspace{2pt} 
    \end{minipage}
    \\ \hline
  \end{tabular} 
   \caption{Example clips from each disagreement quadrant with transcript and labels.}
  \label{tab:quadrant_examples}
\end{table*}